\pdfoutput=1
\documentclass[11pt]{article}

\usepackage{acl}

\usepackage{times}
\usepackage{latexsym}

\usepackage{stfloats}
\usepackage{ulem}
\usepackage{microtype}
\usepackage{makecell}
\usepackage{multirow}
\usepackage{pifont}
\usepackage{comment}
\usepackage{bm}
\usepackage{pifont}
\usepackage{amssymb}
\usepackage{graphicx}
\usepackage{subcaption}
\usepackage{amsthm}
\usepackage{booktabs}

\usepackage{CJK}
\usepackage{wrapfig}
\usepackage{array}
\usepackage{amsmath}

\title{TripleRE: Knowledge Graph Embeddings via Tripled Relation Vectors}

\author{Long Yu\dag, Zhicong Luo\ddag, Huanyong Liu\dag, Deng Lin\dag, Hongzhu Li\dag, Yafeng Deng\dag \\
  Qihoo360, China\dag \\
  Xi'an Jiaotong University, China\ddag \\
  \texttt{\{yulong,luozhicong,lihongzhu,lindeng,liuhuanyong,dengyafeng\}@360.cn} \\ 
  \texttt{zhicong\_luo@stu.xjtu.edu.cn} \\ 
  }

\begin{document}
\begin{sloppypar}
\maketitle
\begin{abstract}
Translation-based knowledge graph embedding has been one of the most important branches for knowledge representation learning since TransE came out. Although many translation-based approaches have achieved some progress in recent years, the performance was still unsatisfactory. This paper proposes a novel knowledge graph embedding method named TripleRE with two versions. The first version of TripleRE creatively divide the relationship vector into three parts. The second version takes advantage of the concept of residual and achieves better performance. In addition, attempts on using NodePiece to encode entities achieved promising results in reducing the parametric size, and solved the problems of scalability. Experiments show that our approach achieved state-of-the-art performance on the large-scale knowledge graph dataset \textit{ogbl-wikikg2}, and competitive performance on other datasets. 
\end{abstract}

\section{Introduction}

Knowledge graphs store huge amounts of structured data in the form of triples, with projects such as WordNet \cite{miller1995wordnet}, Freebase  \cite{bollacker2008freebase}, YAGO \cite{suchanek2007yago} and DBpedia \cite{lehmann2015dbpedia}.
They have gained widespread attraction from their successful use in tasks such as question answering \cite{bordes2014open}, semantic parsing \cite{berant2013semantic}, and named entity disambiguation \cite{zheng2012entity} and so on.

In recent years, knowledge representation is one of the hottest topics in natural language processing as it can be applied to various tasks based on knowledge graphs, including semantic parsing, named entity disambiguation, question answering, etc. A knowledge graph contains finite number of facts represented by triplets. A triplet (h, r, t) contains three components, where h, r, t represent head entities, relations and tail entities, respectively. Knowledge representation is one of the most essential parts in research of knowledge graphs. The mission for knowledge representation is to leverage an appropriate mechanism to represent three components of triplets in numeric vectors. An ideal mechanism for representing knowledge is capable to map the components of triplets to the pre-defined vector space, such as Euclidean space, along with accuracy and some degrees of uncertainty. 

The process of knowledge representation learning could be simply summarized into two phases: the encoding phase and the decoding phase. (i) \textbf{The encoding phase}: The purpose of the encoding phase is to set the embeddings of three components. (ii) \textbf{The decoding phase}: This phase usually includes scoring functions, which are equations include three components, to measure the plausibility of triplets. In other words, the decoding phase is expected to discriminate factual triplets (positive samples) and noisy triplets (negative samples) which are generated according to a selected sampling strategy.

Although many distinctive signs of progress have been achieved in recent years, from our perspectives, one of the most inspired attempts is the blending of information from the structure itself and the logical information embedded in the structure. Knowledge graphs can be viewed as a special kind of directed graphs, where nodes represent entities and edges represent relations. Based on the properties of directed graphs, structural information can be extracted by the positional information of entities in knowledge graphs. The positional information of entities generally represents centrality of the entity nodes and the distance to other nodes. Furthermore, the logical information embedded in the structure can be extracted by the pairs of connected nodes, where the connections indicate the correlation of entities, and edges and relations have equivalence.

Previously, most researches leveraged only the information from the structure of triplets and the logical information embedded in the structure in triplets to learn the knowledge representation. TransE \cite{bordes2013translating} is the fundamental work of translation-based models. Inspired by TransE \cite{bordes2013translating}, many variants, including TransH\cite{wang2014knowledge}, TransR\cite{lin2015learning}, TransAt \cite{qian2018translating}, RotatE \cite{sun2019rotate} and PairRE \cite{chao2020pairre}, handled some issues of TransE\cite{bordes2013translating}, and improved the representing capacities. Translation-based models are reputational in semantics translation. However, all those approaches do not leverage any diversified information of graph structures in either the encoding phase or the decoding phase. 

Structural information can be leveraged in knowledge representation by two different strategies. The first strategy is to extract structural information in the encoding phase. NodePiece \cite{galkin2021nodepiece} introduces an anchor-based approach to fix the size of entity vocabulary. Every entity can be represented by combining of a set of anchors – selected entities. The second strategy is to extract structural information in both the encoding scheme and the decoding scheme. Approaches under this strategy are highly dependent on sub-graph sampling, including GraIL \cite{teru2020inductive} and RedGNN \cite{zhangknowledge}, are usually based on Graph Neural Networks (GNN) \cite{kipf2016semi}, and follow message passing scheme \cite{gilmer2017neural}. However, in addition to training complexity, the interpretability of inference is also unsatisfactory. Hence, from our perspectives, the first strategy is more valuable as it reduces the parametric size, and solves the problems of scalability.

Our approach is inspired by NodePiece \cite{galkin2021nodepiece}, TransAt  \cite{qian2018translating} and pairRE \cite{chao2020pairre}. Our approach chooses to take advantage of NodePiece \cite{galkin2021nodepiece} in the encoding phase as it can significantly reduce the parameters of models. Then, we introduce a novel decoding method, which is inspired by the sub-relations from PairRE \cite{chao2020pairre} and the intuitions from TransAt \cite{qian2018translating}. TransAt \cite{qian2018translating} believes when humans consider triple link prediction tasks, they usually make decisions based on context, not just single node and relation. We argue that the scoring function of PairRE \cite{chao2020pairre} is not complete, and do not follow the hypothesis of translation-based models: $\textbf{h} + \textbf{r}  \approx  \textbf{t}$. Therefore, we introduce a new decoding phase (scoring function) by adding one more sub-relation term. Since our approach uses three sub-relation terms to build the semantics of entire relations, we followed the naming logic of PairRE \cite{chao2020pairre} and named our method \textbf{TripleRE}.

Our approach achieved a competitive performance in large scale datasets - ogbl-wikikg2 and ogbl-biokg \cite{hu2020open}. On the benchmark ogbl-wikikg2, our method achieved the best performance. Compared with other methods, the amount of parameters is significantly reduced. In addition, experiments show that our work also has better results on small datasets such as FB15k \cite{bordes2013translating} and FB15k237 \cite{toutanova2015observed}.

\section{Related work}

This section is going to review several approaches to knowledge representation.
\label{sec.related.work}

\textbf{1) Translation distance models}: 
Based on the translation invariance of the word embedding they found,  TransE \cite{bordes2013translating} expresses the relation as the translation of the entity vectors in $h + r \approx t$, which is the ground-breaking work in the translation distance model. TransE \cite{bordes2013translating} has the advantages of simplicity and scalability, but it cannot model complex relations and multiple relations patterns, such as symmetric relations. The core of a translation distance model is the scoring function. A better scoring function will have a better performance in modeling complex relations such as 1-N, N-1, N-N and relation patterns, such as symmetric/non-relations, inverse relations, combination relations, and sub-relations. TransX series (TransE \cite{bordes2013translating}, TransH \cite{wang2014knowledge}, TransD \cite{ji2015knowledge}, TransR \cite{lin2015learning}) have made up for the shortcomings of TransE's inability to express complex relationships and symmetrical relationships, but they still have many flaws, including the complexity of models and the insufficient expression of the relations. TransAt \cite{qian2018translating} describes the implementation of an attention mechanism for prediction by focusing on only a part of the entity's dimensions. RotatE \cite{sun2019rotate} takes inspiration from Euler's formula and uses the rotation of the vector to express relations. At the same time, it uses complex embedding to model inverse relations. PairRE \cite{chao2020pairre} uses two-stage vectors to express the relations. It is capable of modeling more complex relations, such as sub-relations.

\textbf{2) Bilinear models}: 
The bilinear model is also known as the semantic matching model. For most bilinear models, their core is also the scoring function. Their score functions are based on similarity scores to measure the possibility of the triplets by matching the latent semantics of entities and relations in the embedding vector space. RESCAL \cite{nickel2011three} uses vectors to represent the embedding of the head entity and the tail entity. The relation is expressed as a matrix to model the interaction of the triple. DistMult \cite{yang2014embedding} imposes constraints on the relation matrix, and simplifies the calculation. ComplEX \cite{trouillon2016complex} embeds the head entities, tail entities, and relations into the complex space so that it can better model the antisymmetric relations. HolE \cite{nickel2016holographic} combines the expressiveness of RESCAL \cite{nickel2011three} with the efficiency of DistMult \cite{yang2014embedding}, and defines an operation called circular correlation to calculate the rationality score of fact triples. This calculation can be regarded as compressing the interaction between paired entities so that more interactive information can be captured with less calculation. Circular correlation does not satisfy the exchange law, and HolE \cite{nickel2016holographic} can model asymmetric relations. There is also a bilinear model based on AutoML. AutoSF \cite{zhang2020autosf} uses a determined search algorithm to search the score function with the best performance in the search space. It first learns a set of the converged model parameters on the training set and then searches for a better scoring function on the validation set. Compared to the translation distance models, the bilinear models have shortcomings in encoding relational patterns and leveraging logical information.

\textbf{3) Neural network-based models}: 
With the development of deep learning and the popularity of neural networks, there are many knowledge representation models based on neural networks. ConvE \cite{dettmers2018convolutional} and R-GCN \cite{schlichtkrull2018modeling} can show high performance under desirable parameters, but they are also challenging to manipulate and analyze due to the many parameters and difficulty in explain.

% \textbf{4) sub-graph models}

\textbf{4) NodePiece}: 
NodePiece \cite{galkin2021nodepiece} has a remarkable ability to access structural information in the knowledge graph. By filtering the anchors and encoding other nodes, the model can learn the relative position and centrality of each node in the graph. Anchor entity nodes and context relations represent each entity node, and a vocabulary is constructed for model training, which dramatically reduces the number of parameters and improves the effectiveness of the model. In addition, NodePiece has excellent cooperation with other models that leverage the logical information in a knowledge graph.

\section{Methodology}

The intuition of our approach is similar to the intuition of TransAt \cite{qian2018translating}, which tends to focus on a subset of features for link predictions according to the hierarchical routine of human cognition. Suppose there is a triplet $<$ Yao Ming, married with, $?>$, Yao Ming has multiple identities: a retired basketball player, the chairman of the Chinese Basketball Association, a husband, etc. To identify who is Yao Ming's wife, it is supposed to pay more attention to Yao Ming's identity on being a husband. Therefore, we need a mechanism to concentrate on Yao Ming's features related to being a husband, and reduce attention to irrelevant features. TransAt \cite{qian2018translating} follows PCA \cite{dunteman1989principal}; it argues that if a feature is related to a relation r, then the variance of this feature between candidate entities and r should be large, and vice versa. Therefore, TransAt \cite{qian2018translating} sets a threshold to determine whether to retain or drop out a feature from the whole feature set. In other words, TransAt \cite{qian2018translating} leverages a subset of all features. Inspired by PairRE \cite{chao2020pairre}, every relation is represented as the projections of the node, and entities are mapped to different feature spaces by performing the dot product with the relation, which is equivalent to feature filtering. After mapping entities to a specific feature space with the involvements of relations, the decision space becomes smaller, which could lead to better predictions.% An experiment can be added here to support

\textbf{TripleREv1}: Two versions of TripleRE are suggested in this paper. For the first version, we split the relations into three segments, $r_h$, $r_m$, and $r_t$. $r_h$ and $r_t$ are the projected parts of the relation for the head entities and tail entities respectively, and $r_m$ is the translation part of the relation. $\circ$ represents the Hadamard product. The scoring function used in our decoding phase is as follows:
\begin{equation}\label{d}
f_r(h, t) = -\lVert h \circ r^{H} - t \circ r^{T} + r^{M} \lVert
\end{equation}
\noindent

where $r^{H} = r_{h}$, $r^{T} = r_{t}$ and $r^{M} = r_{m}$.
%\usepackage{caption}
% \usepackage{subfigure}
%导言区域要添加以上两个包
\begin{figure*}[h]
   \begin{subfigure}{.33\textwidth}
    \centering
    \includegraphics[width=0.8\linewidth]{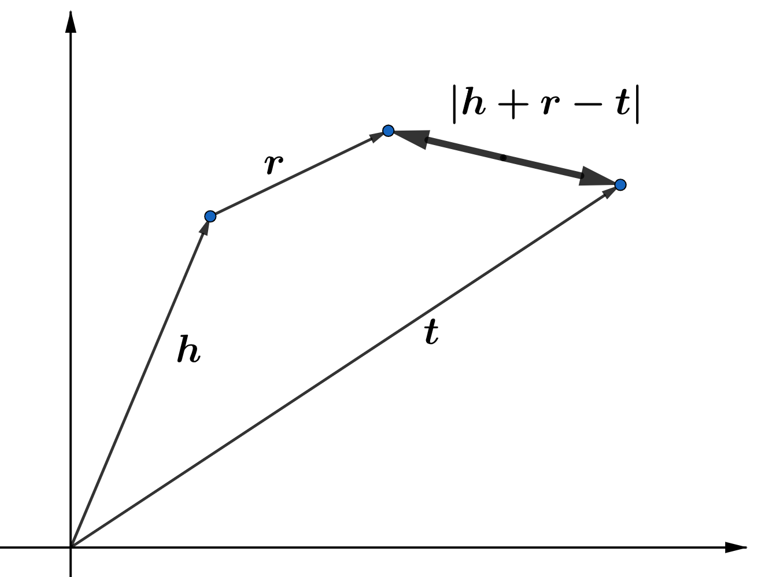}
    \caption{TransE}
    \label{fig:transe}
  \end{subfigure}
  \begin{subfigure}{.33\textwidth}
    \centering
    \includegraphics[width=0.85\linewidth]{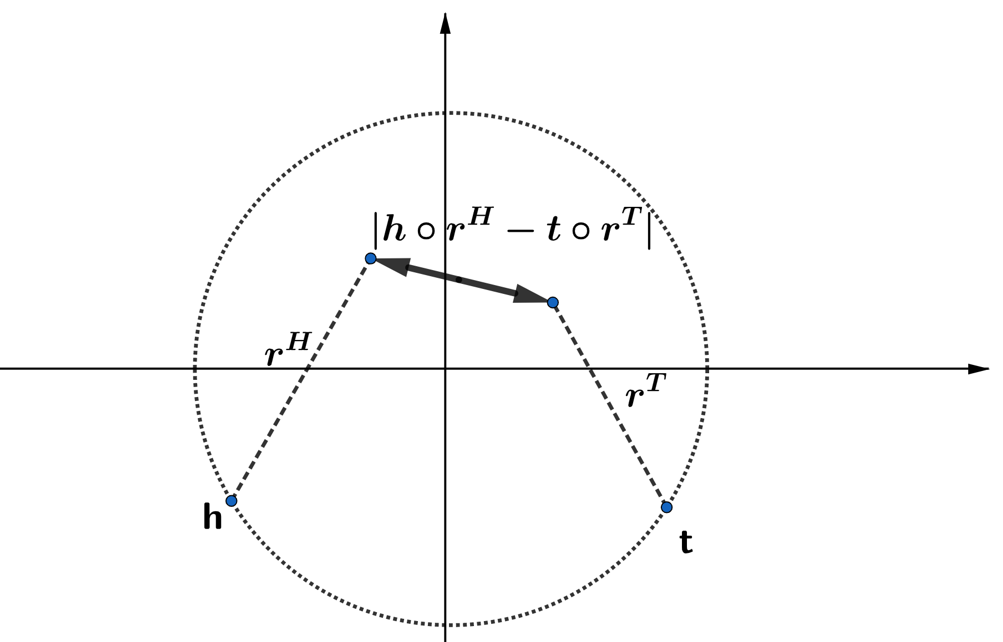}
    \caption{PairRE}
    \label{fig:rotate}
  \end{subfigure}
  \begin{subfigure}{.33\textwidth}
    \centering
    \includegraphics[width=0.85\linewidth]{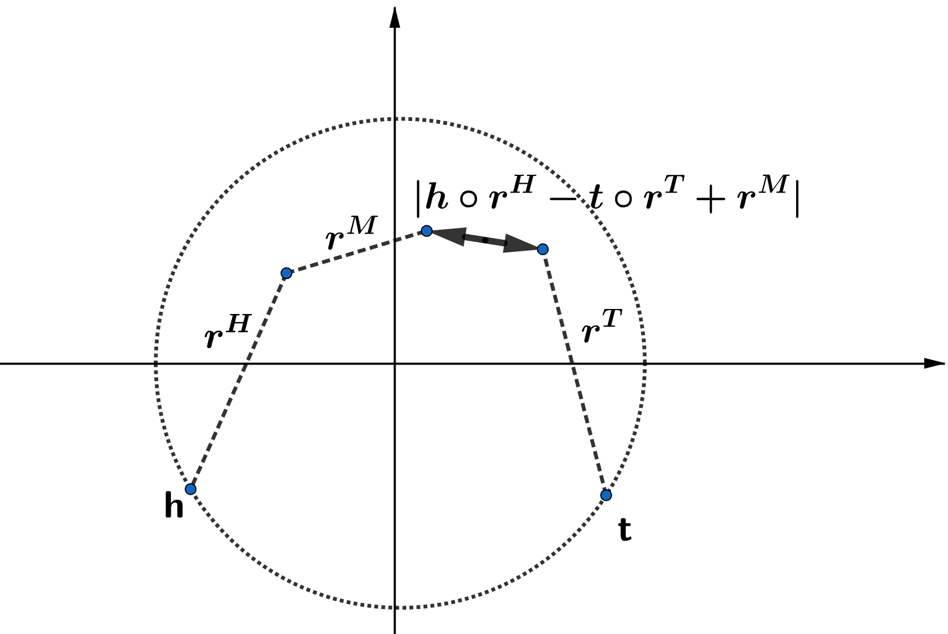}
    \caption{TripleRE}
    \label{fig:pairre}
  \end{subfigure}
\caption{Illustration of TransE, PairRE and TripleRE. For TripleRE, all entities are on the unit circle. $r^{H}$ and $r^{T}$ project entities to different locations, $r^{M}$ is response for the translation part.
}
\label{fig:Illustration}
\end{figure*}

% \vspace{-0.5cm}
% \begin{figure}
% \vspace{-0.3cm}
% \caption{Illustration of TransE, PairRE and TripleRE. For TripleRE, all entities are on the unit circle. $r^{H}$ and $r^{T}$ project entities to different locations, $r^{M}$ is response for the translation part. } \label{fig1}
% \end{figure}

The scoring function of TripleREv1 is similar to the scoring function of PairRE\cite{chao2020pairre}. However, according to Figure~\ref{fig:1}, PairRE\cite{chao2020pairre} is essentially a special case of TripleREv1.

% An experiment can be added here to support residual thought
\textbf{TripleREv2}: TripleREv2 is a better version optimized based on TripleREv1. We believe that the segments of relations $r_{h}$ and $r_{t}$  play different roles in the decoding phase. For learning better representations of $r_{h}$ and $r_{t}$,  TripleREv2 takes advantage of the concept of residual. The original model needs to learn $r^{H}=r_{h}$, $r^{T}=r_{t}$, we made the following changes: $r^{H}=r_{h} + u*e$, $r^{T}=r_{t} + u*e$, The candidate value of $u$ is heuristically chosen, and we took 1, 0.5, 0.25 and 0.125 for u in our experiments, taking the best result as the value for $u$. Now the model can be regarded as using TransE\cite{bordes2013translating} as the baseline, using $r^{H}$ and $r^{T}$ to learn the transformation of the residuals. Experiments proved that the involvement of residual mechanism does improve the performance. The scoring function of TripleREv2 is as follows:
\noindent

\begin{equation}\label{d}
\begin{aligned}
f_r(h, t) = -\lVert h \circ (r_{h} + u*e) - 
\\t \circ (r_{t} + u*e) + r_{m} \lVert
\end{aligned}
\end{equation}

Two parameters $u$ and $e$ are added to the scoring function, where $u$ is a constant and set default to 1, and $e$ is a unit vector.

\textbf{TripleRE+NodePiece}: NodePiece \cite{galkin2021nodepiece} is an anchor-based method for learning a fixed-size vocabulary for any connected multi-relational graph. In NodePiece \cite{galkin2021nodepiece}, an atomic set consists of anchors and all relation types, which allow a combined number of sequences to be built from a limited vocabulary of atoms. NodePiece \cite{galkin2021nodepiece} has several advantages. First, NodePiece\cite{galkin2021nodepiece} can significantly reduce the number of parameters; second, NodePiece \cite{galkin2021nodepiece} can extract certain sub-graph features; third, NodePiece \cite{galkin2021nodepiece} can be used in the encoding phase to connect with other score functions seamlessly. We use NodePiece \cite{galkin2021nodepiece} in the encoding phase and TripleRE in the decoding phase.

We followed the NodePiece \cite{galkin2021nodepiece} ablation experiment and used three different methods to encode the nodes. First, nodes are represented by K-nearest anchors and the translation part information of TripleRE; second, K-nearest anchors represent nodes, and third, nodes are represented by K-nearest anchors and full relation information.

\section{Experiments}

\subsection{Datasets}

Five datasets on different scales were selected for experiments to verify the effectiveness of the proposed approaches.Especially, results in FB15k and WN18 are only to be compared to TransAT and PairRE.

\begin{itemize}

\item \noindent\textbf{ogbl-wikikg2}: ogbl-wikikg2 \cite{hu2020open} is a large-scale knowledge graph extracted from Wikidata knowledge base \cite{vrandevcic2014wikidata}, and contains 2,500,604 of entities and 535 relation types.

\item \noindent\textbf{ogbl-biokg}: ogbl-biokg \cite{hu2020open} is a knowledge graph in fields of biomedicine, and extracted from biomedical datasets. It contains 93,773 entities, which can be classified into five categories: diseases, proteins, drugs, side effects and protein functions, and 51 relation types. 

\item \noindent\textbf{FB15k}: FB15k \cite{bordes2013translating} is a small-scale knowledge graph, which contains 14,951 entities and 1,345 relation types. 

\item \noindent\textbf{FB15k237}: FB15k237 \cite{toutanova2015observed}, a subset of FB15k, contains the same number of entities with FB15k but 237 relation types as all inverse relations are removed from FB15k.

\item \noindent\textbf{WN18}: WN18 \cite{bordes2013translating} is a small-scale knowledge graph, which contains 40,943 entities and 18 relation types

% \item \noindent\textbf{WN18RR}: WN18RR \cite{dettmers2018convolutional}, a subset of WN18, removes all the inverse relations of WN18. It contains the same number of entities with WN18 but 11 relation types.

\end{itemize}

\subsection{Implemention}

To verify the performance of our model, we conduct comparative experiments on large-scale and small-scale datasets, respectively. Considering our work is similar to the idea of TransAt \cite{qian2018translating}, we compare with TransAt \cite{qian2018translating} on the FB15k and WN18 datasets. Details are shown in Table~\ref{tab:tlp1}. In the implementation process, results of TransAt \cite{qian2018translating} were completely trained and predicted by using its open-source code. TripleRE was also limited to train and predict in the same dimension for a fair comparison.

\begin{table}[t]
\begin{center}
\resizebox{0.48\textwidth}{!}{
\begin{tabular}{c|ccc|ccc}
\hline
- &\multicolumn{3}{c|}{WN18} & \multicolumn{3}{c}{FB15k} \\ \hline
\textbf{Model} &$\#$MRR &Hit@10 &Dim &$\#$MRR &Hit@10 &Dim  \\ \hline
TransAt &64.46 &95.10 &100 &50.77 &78 &200 \\
TripleRE &77.25 &95.41 &100 &67.53 &83.03 &200 \\ \midrule
\end{tabular}
}
\end{center}
\caption{Comparison results of TransAt and TripleRE under FB15k and WN18.We report the performance in a "filtered" setting as in TransAt on test datasets}
\label{tab:tlp1}
\end{table}

PairRE\cite{chao2020pairre} conducts comparative experiments on two small datasets, FB15k and FB15k237. We trained and predicted on both datasets using baseline's hyperparameters for comparison. The specific results are shown in Table~\ref{tab:tlp2}, and the results of other models are taken from PairRE.

\begin{table*}[h]
%\newcolumntype{C}{ @{}>{${}}c<{{}$}@{} }
\begin{center}
\resizebox{0.9\textwidth}{!}{
\begin{tabular}{c|ccccc|ccccc}
\hline
- &\multicolumn{5}{c|}{FB15k} & \multicolumn{5}{c}{FB15k-237} \\ \hline
\textbf{Model} & MR & MRR &Hit@10 &Hit@3 &Hit@1 &MR &MRR &Hit@10 &Hit@3 &Hit@1 \\ \hline
TransE & - & 0.463 & 0.749 & 0.578 & 0.297 & 357 & 0.294 & 0.465 & - & -\\
DistMult & 42 & 0.798 & \uline{0.893} & - & - & 254 & 0.241 & 0.419 & 0.263 & 0.155\\
HoIE & - & 0.524 & 0.739 & 0.759 & 0.599 & - & - & - & - & -\\
ConvE & 51 & 0.657 & 0.831 & 0.723 & 0.558 & 244 & 0.325 & 0.501 & 0.356 & 0.237 \\
ComplEx & - & 0.692 & 0.840 & 0.759 & 0.599 & 339 & 0.247 & 0.428 & 0.275 & 0.158 \\
SimpIE \cite{kazemi2018simple} & - & 0.727 & 0.838 & 0.773 & 0.66 & - & - & - & - & - \\
RotatE & 40 & 0.797 & 0.884 & 0.830 & 0.746 & 177 & 0.338 & 0.533 & 0.375 & 0.241 \\
Seek \cite{xu2020seek} & - & \textbf{0.825} & 0.886 & \uline{0.841} & \textbf{0.792} & - & - & - & - & - \\
OTE \cite{tang2019orthogonal} & - & - & - & - & - & - & \uline{0.351} & 0.537 & 0.388 & \uline{0.258} \\
GC-OTE \cite{tang2019orthogonal} & - & - & - & - & - & - & \textbf{0.361} & \uline{0.550} & \textbf{0.396} & \textbf{0.267} \\
PairRE & \uline{37.7} & \uline{0.811} & \textbf{0.896} & \textbf{0.845} & \uline{0.765} & \uline{160} & \uline{0.351} & 0.544 & 0.387 & 0.256 \\
TripleRE & \textbf{35.7} & 0.747 & 0.877 & 0.813 & 0.662 & \textbf{142} & \uline{0.351} & \textbf{0.552} & \uline{0.392} & 0.251 \\
\midrule
\end{tabular}
}
\caption{Comparison of TripleRE and other models on FB15k and FB15k-237. Bold numbers indicate best performance, underlined indicates next best.}
\label{tab:tlp2}
\end{center}
\end{table*}

\begin{table*}[h]
\begin{center}
\resizebox{0.9\textwidth}{!}{
\begin{tabular}{c|ccc|ccc}
\hline
- &\multicolumn{3}{c|}{ogbl-wikikg2} & \multicolumn{3}{c}{ogbl-biokg} \\ \hline
Model & Test MRR & Valid MRR & \#P & Test MRR & Valid MRR & \#P \\ \midrule
AutoSF & 0.5458 ± 0.0052 & 0.5510 ± 0.0063 & 500M & \textbf{0.8309 ± 0.0008} & \textbf{0.8317 ± 0.0007} & 93M \\
PairRE & 0.5208 ± 0.0027 & 0.5423 ± 0.0020 & 500M & 0.8164 ± 0.0005 & 0.8172 ± 0.0005 & 187M \\
RotatE & 0.4332 ± 0.0025 & 0.4353 ± 0.0028 & 1,250M & 0.7989 ± 0.0004 & 0.7997 ± 0.0002 & 187M \\
TransE & 0.4256 ± 0.0030 & 0.4272 ± 0.0030 & 1,250M & 0.7452 ± 0.0004 & 0.7456 ± 0.0003 & 187M \\
ComplEx & 0.4027 ± 0.0027 & 0.3759 ± 0.0016 & 1,250M & 0.8095 ± 0.0007 & 0.8105 ± 0.0001 & 187M \\
DistMult & 0.3729 ± 0.0045 & 0.3506 ± 0.0042 & 1,250M & 0.8043 ± 0.0003 & 0.8055 ± 0.0003 & 187M \\
TripleRE & \uline{0.5794 ± 0.0020} & \uline{0.6045 ± 0.0024} & 500M & 0.8191 ± 0.0014 & 0.8192 ± 0.0008 & 187M \\
TripleREv2 & \textbf{0.6045 ± 0.0017}  & \textbf{0.6117 ± 0.0008} & 500M & \uline{0.8272 ± 0.0007} & \uline{0.8281 ± 0.0006} & 187M \\ 
\midrule
\end{tabular}
}
\caption{Comparison of the results of ogbl-wikikg2 and ogbl-biokg. Bold numbers indicate best performance, underlined indicates next best.\#P is a total parameter count (millions).}
\label{tab:tlp3}
\end{center}
\end{table*}

\begin{table*}[!htb]
%\newcolumntype{C}{ @{}>{${}}c<{{}$}@{} }
\begin{center}
\resizebox{0.9\textwidth}{!}{
\begin{tabular}{@{}lccc@{}}
 & \multicolumn{3}{c}{ogbl-wikikg2} \\ \cmidrule(l){2-4} 
Model & Test MRR & Valid MRR & \#P \\ \midrule
ComplEx-RP \cite{chen2021relation} (50dim) & \uline{0.6392 ± 0.0045} & \uline{0.6561 ± 0.0070} & 250,167,400 \\
TripleRE & 0.5794 ± 0.0020 & 0.6045 ± 0.0024 & 500,763,337 \\
NodePiece + AutoSF & 0.5703 ± 0.0035 & 0.5806 ± 0.0047 & \textbf{6,860,602} \\
AutoSF & 0.5458 ± 0.0052 & 0.5510 ± 0.0063 & 500,227,800 \\
PairRE (200dim) & 0.5208 ± 0.0027 & 0.5423 ± 0.0020 & 500,334,800 \\
RotatE (250dim) & 0.4332 ± 0.0025 & 0.4353 ± 0.0028 & 1,250,435,750 \\
TransE (500dim) & 0.4256 ± 0.0030 & 0.4272 ± 0.0030 & 1,250,569,500 \\
ComplEx (250dim) & 0.4027 ± 0.0027 & 0.3759 ± 0.0016 & 1,250,569,500 \\
DistMult (500dim) & 0.3729 ± 0.0045 & 0.3506 ± 0.0042 & 1,250,569,500 \\
TripleRE + NodePiece & \textbf{0.6582 ± 0.0020} & \textbf{0.6616 ± 0.0018} & \uline{7,289,002} \\
\midrule
%\multicolumn{1}{r}{random anchors} & 2M & 0.257 & 0.420 & 79 &  & 0.393 & 0.511 & \\
\end{tabular}
}
\caption{For TripleRE+NodePiece, We compared with all other models. Bold numbers indicate best performance, underlined indicates next best.}
\label{tab:tlp4}
\end{center}
\end{table*}

\begin{table}[!htb]
%\newcolumntype{C}{ @{}>{${}}c<{{}$}@{} }
\begin{center}
\resizebox{0.48\textwidth}{!}{
\begin{tabular}{@{}lccc@{}}
 & \multicolumn{3}{c}{ogbl-wikikg2} \\ \cmidrule(l){2-4} 
Model & Test MRR & Valid MRR & \#P \\ \midrule
v1 + anchor + r & 0.6262 ± 0.0015 & 0.6296 ± 0.0010 & 7,289,002 \\
v1 + anchor + $rm$ & 0.6331 ± 0.0014 & 0.6422 ± 0.0020 & 7,289,002 \\
v1 + anchor & 0.6477 ± 0.0016 & 0.6527 ± 0.0013 & 6,329,002 \\
v2 + anchor + r & 0.6573 ± 0.0014 & 0.6658 ± 0.0009 & 7,289,002 \\
v2 + anchor + $rm$ & 0.6582 ± 0.0020 & 0.6616 ± 0.0018 & 7,289,002 \\
v2 + anchor & \textbf{0.6627 ± 0.0018} & \textbf{0.6745 ± 0.0015} & \textbf{6,329,002} \\
\midrule
%\multicolumn{1}{r}{random anchors} & 2M & 0.257 & 0.420 & 79 &  & 0.393 & 0.511 & \\
\end{tabular}
}
\caption{TripleREv1 and TripleREv2 respectively conduct three different encoding experiments. "anchor + r" means nodes are represented by K-nearest anchors and full relation information. "anchor + $rm$" means nodes are represented by K-nearest anchors and the translation part information of TripleRE. "anchor" means nodes are represented by K-nearest anchors. }
\label{tab:tlp5}
\end{center}
\end{table}

For TripleRE, we selected the results of a single model in the large-scale datasets ogbl-wikikg2 and ogbl-biokg for comparison. The details are shown in Table~\ref{tab:tlp3}. In ogbl-wikikg2 and ogbl-biokg, we increased the max steps and reduced the lr to 0.0005; other hyperparameters are consistent with the baseline. For TripleREv2, another hyperparameter u needs to be adjusted compared to the baseline. We set u to 1 in ogbl-wikikg2 and set u to 0.125 in ogbl-biokg.

For TripleRE+NodePiece, since NodePiece is not suitable for the data set of biokg\cite{galkin2021nodepiece}, we compared it on the large-scale data set ogbl-wikikg2, and the result shows on Table~\ref{tab:tlp4}. For the implementation of TripleRE + NodePiece, we followed the work of NodePiece+AutoSF, and the hyperparameters are the same as NodePiece+AutoSF, except for max steps. Detailed results are shown in Table~\ref{tab:tlp4}.

We followed NodePiece's ablation experiments and conducted experiments on large-scale datasets ogbl-wikikg2. We tried three different node encoding methods. The first we call "anchor + rm"; the second we call "anchor"; the third we call "anchor + r." Detailed results are shown in Table~\ref{tab:tlp5}.

The above experiments in ogbl-wikikg2 and ogbl-biokg were repeated ten times, with random numbers set from 0 to 9.

\section{Main Results}

As can be seen from Tables~\ref{tab:tlp3} and ~\ref{tab:tlp4}, TripleRE has significantly improved the performance on the large-scale datasets ogbl-wikikg2 and ogbl-biokg. On ogbl-wikikg2, TripleRE+NodePiece achieved state-of-the-art performance, and parametric size was reduced by more than three times compared with the approach in second place.

From Table~\ref{tab:tlp1}, we can see that on the small datasets FB15k and WN18, our work has an obvious advantage over TransAt \cite{qian2018translating}. Table~\ref{tab:tlp2} shows the comparison between TripleRE and other models. The results show that our model has obtained the best performance in MR and achieved competitive results in other metrics. 
% This indicates TripleRE improves precision on small datasets but reduces recall.

Table~\ref{tab:tlp5} shows the various experiments we performed at TripleRE+NodePiece. We conduct three different experiments on encoding node representations for TripleREv1 and TripleREv2. The experiments in Table~\ref{tab:tlp5} show that the "anchor" method works best on TripleRE+NodePiece, both Test MRR and Valid MRR achieve the best performance, and the number of parameters is also less than the lowest model (NodePiece+AutoSF), which means we have achieved the state-of-the-art with the least amount of parameters on wikikg2.
% First,  Second, anchor is the best solution, which is also consistent with Nodepiece's conclusion. Third, the TripleRE+Nodepiece method is better than the single TripleRE model.

\section{Conclusion and future work}

The contributions of this paper can be summarized as follows:

\begin{itemize}

    \item \noindent This paper proposes a novel approach, TripleRE, in the decoding phase of knowledge representation learning. As a translation-based approach, TripleRE is advantageous in efficiency (in both training and inferencing stages) and simplicity.
    
    \item \noindent In TripleREv2, it utilizes the residual mechanism to optimize knowledge representation. Considering residual mechanism has made a tremendous difference in many fields of deep learning, it is also promising to apply it to knowledge representation.
    
    \item \noindent TripleRE achieved competitive performance on large-scale datasets: \textit{ogbl-wikikg2} and \textit{ogbl-biokg}.

    \item \noindent  The combination of NodePiece and TripleRE achieved state-of-the-art performance in the benchmark \textit{ogbl-wikikg2}.

\end{itemize}

The future work can be summarized in two directions. The first direction is to explore the methods to leverage structural information. One of the potential proposals is to improve the anchor selection method of NodePiece. Another direction is to conduct more experiments on other large-scale knowledge graphs to test the capacities of TripleRE and TripleRE+NodePiece further.

\paragraph*{Supplemental Material Statement:} Source code for transAt in Section 4 is available from Github\footnote{https://github.com/ZJULearning/TransAt}. The experimental source code at FB15k and FB15k-237 is available from Github\footnote{https://github.com/DeepGraphLearning/KnowledgeGraph- Embedding}. Source code for TripleRE and TripleRE+NodePiece will be submitted via easy chair. After the review, we will open source the code on Github.

\normalem
\bibliography{anthology,custom}
\bibliographystyle{acl_natbib}

\appendix
\end{sloppypar}
\end{document}